\documentclass{article}

\PassOptionsToPackage{numbers, compress}{natbib}

\usepackage[preprint]{neurips_2026}

\usepackage[utf8]{inputenc}
\usepackage[T1]{fontenc}
\usepackage{hyperref}
\hypersetup{
    colorlinks=true,
    urlcolor=black,
    linkcolor=black,
    citecolor=black
}
\usepackage{url}
\usepackage{booktabs}
\usepackage{amsfonts}
\usepackage{nicefrac}
\usepackage{microtype}
\usepackage{multirow}
\usepackage{graphicx}
\usepackage{dblfloatfix}
\usepackage{caption}
\usepackage{float}
\usepackage{enumitem}
\usepackage[table]{xcolor}
\usepackage[table]{xcolor}

\definecolor{nvidiagreen}{HTML}{76B900}

\DeclareRobustCommand{\iconhref}[3]{%
  \texorpdfstring{%
    \href{#1}{%
      \raisebox{-0.15em}{\includegraphics[height=1em]{#2}}\,\textcolor{nvidiagreen}{#3}%
    }%
  }{#3}%
}

\title{Light Interaction: Training-Free Inference Acceleration for Interactive Video World Models}

\author{
  Jiacheng Lu$^{1}$ \quad Haoyi Zhu$^{2}$ \quad Sipei Yi$^{1}$ \quad Enze Xie$^{2}$ \quad Yu Li$^{1}$\thanks{Corresponding author: Yu Li at yu.li.sallylee@gmail.com.} \quad Cheng Zhuo$^{1}$\\
  $^{1}$Zhejiang University \quad $^{2}$NVIDIA
}
\usepackage{amsmath}

\begin{document}

\maketitle

\vspace{-2em}
\begin{center}
{\normalsize
\iconhref{https://2843721358l-del.github.io/Light-Interaction-Project/}{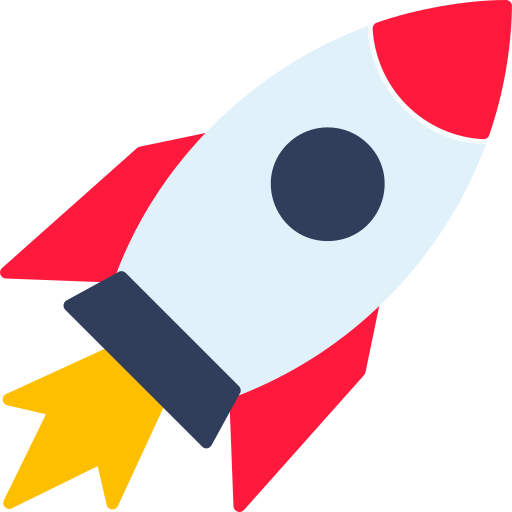}{Project Page}
\quad
\iconhref{https://github.com/2843721358l-del/Light-Interaction-Project}{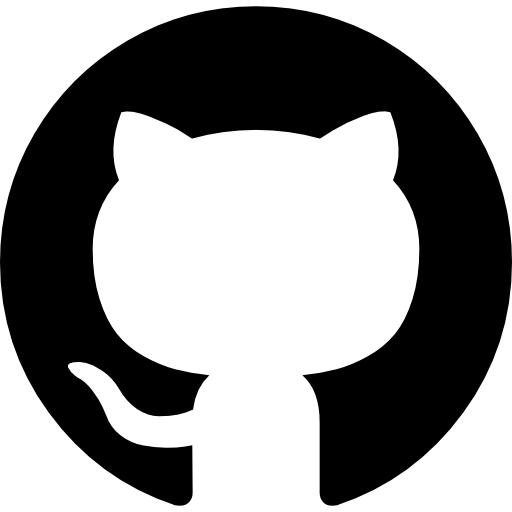}{GitHub}
}
\end{center}
\vspace{0.5em}

\begin{center}
\begin{minipage}{\textwidth}
    \centering
    \normalsize
    \parindent=0pt

    \colorbox{gray!4}{
    \begin{minipage}{\dimexpr\linewidth-10pt\relax}
        \vspace{0.1em}
        \centering
        \normalsize HY-WorldPlay 480P, Image-to-Video \\[0.3em]
        \normalsize Original \ \textcolor{green!80!black}{| Latency = 228.60 s} \\[0.1em]
        \includegraphics[width=\linewidth]{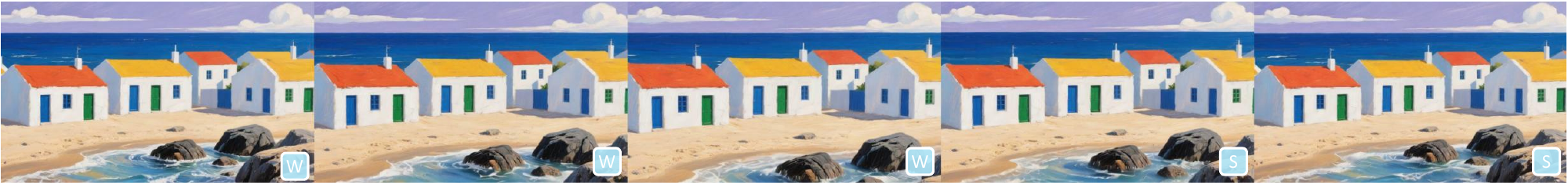} \\[0.1em]

        \normalsize Light Interaction \ \textcolor{red}{| PSNR = 24.81 | Latency = 88.24 s | Speedup 2.59$\times$} \\[0.1em]
        \includegraphics[width=\linewidth]{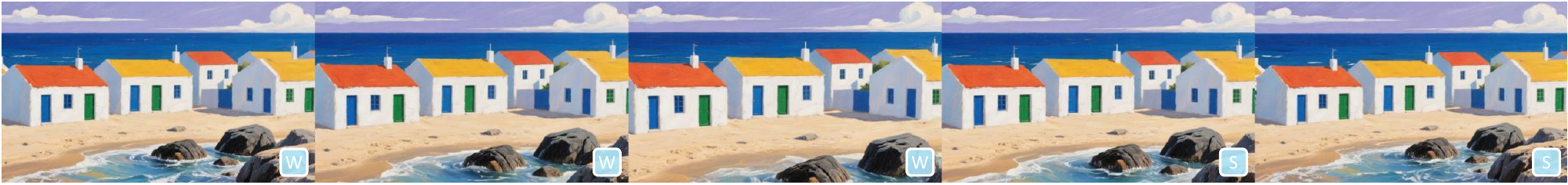} \\[0.1em]

        \parbox{\linewidth}{\scriptsize
        Serene coastal village painting, whitewashed houses with bright orange/yellow roofs, sandy beach, deep blue sea, gentle waves on rocks, purple sky with clouds, oil painting style, soft sunlight, calm and peaceful atmosphere, smooth camera pan, cinematic.
        }
    \end{minipage}
    }

    \vspace{0.3em}

    \colorbox{gray!4}{
    \begin{minipage}{\dimexpr\linewidth-10pt\relax}
        \vspace{0.1em}
        \centering
        \normalsize Matrix-Game-3.0 720P, Image-to-Video \\[0.3em]
        \normalsize Original \ \textcolor{green!80!black}{| Latency = 59.70 s} \\[0.1em]
        \includegraphics[width=\linewidth]{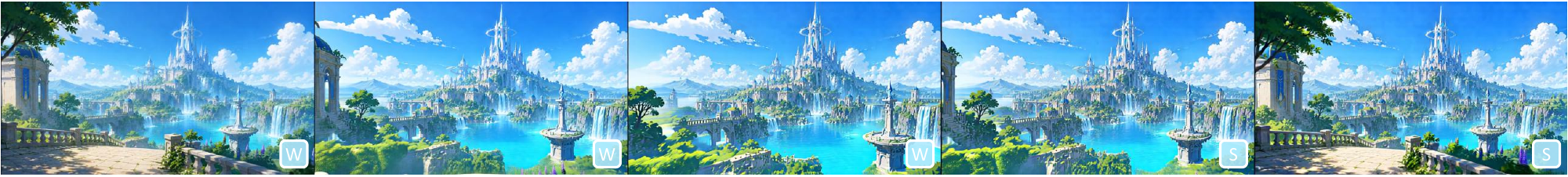} \\[0.1em]

        \normalsize Light Interaction \ \textcolor{red}{| PSNR = 17.76 | Latency = 37.07 s | Speedup 1.61$\times$} \\[0.1em]
        \includegraphics[width=\linewidth]{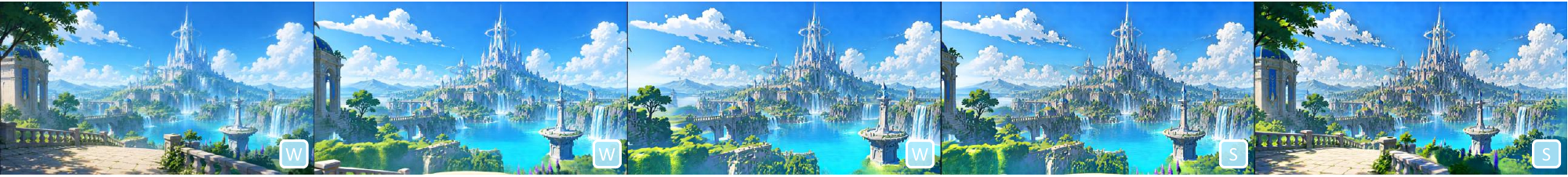} \\[0.1em]

        \parbox{\linewidth}{\scriptsize
        Anime-style fantasy castle, bright blue sky with drifting clouds, cascading waterfalls into a turquoise lake, sunlight filtering through swaying leaves, slow camera dolly-in, soft ambient light, peaceful and magical atmosphere, smooth natural motion, vibrant colors.
        }
    \end{minipage}
    }
    \vspace{-0.5em}

    \captionof{figure}{Light Interaction accelerates interactive video world models while maintaining quality. On a single A100 GPU, it achieves up to \textbf{2.59$\times$ and 1.61$\times$} speedup on HY-WorldPlay and Matrix-Game-3.0, with PSNR 24.81 and 17.76, respectively. (Camera trajectory: Forward, then Backward)
    }
    \label{fig:teaser}
\end{minipage}
\end{center}
\vspace{0.3em}

\begin{abstract}

Interactive video world models generate video chunk by chunk in response to user-controlled camera movements, paving the way toward real-time game simulation, virtual scene navigation, and embodied AI training. However, scaling to long interactive trajectories is prohibitively expensive: generating a 10-second video on HY-WorldPlay with a single A100 GPU can take >200 seconds due to growing context memory, quadratic attention complexity, and repeated denoising steps.
Existing acceleration methods such as cache compression, denoising step reduction, and sparse attention either adopt uniform strategies or fail to deliver practical speedups in autoregressive (AR) settings due to causal constraints and/or asymmetric Q/K lengths.
To address these challenges, we present \textbf{Light Interaction}, a training-free acceleration framework for interactive video world models. Our key insight is that interaction naturally enables adaptive computation: retrieved spatial memory can be discarded during novel scene exploration, temporal windows can shrink under large local latent dynamics, and early-step model outputs can be reused when the camera revisits familiar regions. Based on these observations, we introduce: (1) \emph{adaptive context management} prunes spatial memory by camera-pose-aware similarity and adjusts temporal windows according to local latent dynamics; (2) \emph{denoising cache acceleration} reuses early-step model outputs for intermediate denoising steps in familiar scenes.
Finally, we make sparse attention practical in the AR setting by introducing (3) \emph{hardware-software co-designed sparse attention} which uses Triton fused kernels to close the gap between algorithmic sparsity and realized speedup. Evaluated on HY-WorldPlay and Matrix-Game-3.0, Light Interaction achieves up to \textbf{2.59$\times$} speedup without model retraining, while reaching \textbf{24.81} PSNR against the original model on HY-WorldPlay, maintaining competitive visual quality.

\end{abstract}

\section{Introduction}
\label{sec:intro}

Interactive video world models --- systems in which an agent continuously navigates a dynamically synthesized world --- are becoming increasingly important for game simulation, virtual scene exploration, and embodied AI~\cite{alonso2024diffusion, valevski2024diffusion, bar2025navigation}. Systems such as HY-WorldPlay~\cite{sun2025worldplay} and Matrix-Game-3.0~\cite{li2026matrixgame3} generate video chunk-by-chunk with camera-pose-aware memory retrieval, enabling long-horizon geometric consistency under interactive camera trajectories. However, scaling to long interactive trajectories is prohibitively expensive due to growing context memory, quadratic 3D spatio-temporal attention, and repeated Transformer executions across denoising steps. For example, generating 10 seconds of video on HY-WorldPlay with a single A100 GPU can take over 200 seconds.

Existing acceleration methods only partially address this bottleneck. KV cache compression methods~\cite{xiao2023efficient, zhang2023h2o, liu2024scissorhands} reduce context memory by compressing cached history, but do not determine whether retrieved spatial memory is useful under changing camera trajectories. Denoising cache methods~\cite{ma2024deepcache, zhao2024pab, lv2024fastercache, liu2024timestep, chen2024delta} reuse cached denoising outputs to reduce repeated computation, but do not determine when such reuse is reliable. Sparse attention methods~\cite{zhang2025spargeattn, xi2025sparse, yang2025sparsevideogen2, wu2025vmoba, lv2026lightforcing} reduce theoretical attention cost, but their practical gains are often weakened by causal layout constraints and gather/scatter overhead in AR execution. As a result, existing approaches either apply uniform computation across interaction scenarios or fail to achieve practical acceleration in AR generation.

Our key observation is that interaction naturally enables adaptive computation, meaning that the usefulness of different computation evolves with interaction dynamics. First, pose-aware retrieval similarity can indicate whether long-range retrieved spatial memory remains informative, distinguishing \emph{novel exploration}, where such memory is often unreliable, from \emph{trajectory revisiting}, where historical views become useful again. Second, the utility of temporal context depends on local latent dynamics rather than a fixed history budget. Third, during revisiting, early denoising-step outputs can approximate intermediate steps, reducing repeated Transformer computation.

At the same time, making the remaining attention efficient is another systems challenge. Even after adaptive context management and denoising simplification, autoregressive generation still requires attention over long historical visual memory; without AR-aware layout and fused execution, sparse patterns can lose much of their theoretical benefit to gather/scatter and layout-conversion overhead.

We present \textbf{Light Interaction}, a novel training-free inference acceleration framework for interactive video world models. The core principle is \emph{trajectory-dependent adaptive computing}: Light Interaction exploits pose-aware retrieval similarity to gate retrieved spatial memory and denoising reuse, uses local latent dynamics to adapt temporal context, and employs an AR-aware sparse attention backend to make the remaining attention computation hardware-efficient. Our contributions are as follows.
\begin{itemize}
\item We propose \textbf{adaptive context management}, which disables unreliable spatial memory using camera-pose-aware retrieval similarity and adaptively adjusts the temporal context window according to local latent dynamics.
\item We propose a \textbf{denoising cache acceleration} that reuses early-step model outputs for intermediate denoising steps when camera-pose-aware retrieval similarity indicates reliable revisiting, while preserving the final step for quality correction.
\item We introduce \textbf{hardware-software co-designed 3D block sparse attention}, which preserves text and current-chunk tokens, sparsifies only historical visual KV blocks, and uses fused Triton kernels to remove layout-conversion and gather/scatter overhead under autoregressive causal constraints.
\end{itemize}
Experiments on HY-WorldPlay and Matrix-Game-3.0 --- the two representative open-source interactive video world models --- demonstrate up to \textbf{2.59$\times$} speedup without model retraining.

\section{Related Work}
\label{sec:related_work}

\noindent\textbf{Autoregressive Video Generation and Interactive World Models.} Compared with bidirectional video diffusion models~\cite{wan2025wan, yang2024cogvideox, brooks2024video}, autoregressive generation predicts frames sequentially~\cite{zhang2025packing, huang2025self, gu2025long, henschel2025streamingt2v}, naturally supporting streaming and interactive applications~\cite{alonso2024diffusion, bar2025navigation, valevski2024diffusion}. For long-term spatial consistency, prior work uses explicit 3D reconstruction~\cite{li2025vmem, yu2025wonderworld, ren2025gen3c} or camera-pose-aware retrieval~\cite{xiao2025worldmem, yu2025context}. Recent works such as HY-WorldPlay~\cite{sun2025worldplay} and Matrix-Game-3.0~\cite{li2026matrixgame3} follow the latter paradigm, but primarily use retrieval for consistency preservation rather than inference acceleration.

\noindent\textbf{Context Management.} For retrieved spatial memory, KV cache compression methods~\cite{xiao2023efficient, zhang2023h2o, liu2024scissorhands} evict tokens based on attention scores to bound memory, and Light Forcing~\cite{lv2026lightforcing} applies uniform KV pruning for interactive video generation. For temporal context, autoregressive video models typically use a fixed sliding window~\cite{henschel2025streamingt2v, gu2025long}. These methods use uniform policies regardless of camera trajectory, whereas our method adapts both retrieved spatial memory and temporal context.

\noindent\textbf{Denoising Cache Acceleration.} Step-reduction methods use improved solvers~\cite{song2020denoising, lu2022dpm} or distillation~\cite{salimans2022progressive, yin2024one}; CausVid~\cite{yin2025slow} and Self-Forcing~\cite{huang2025self} make few-step ($K\leq4$) AR inference practical. Caching methods exploit redundancy across denoising timesteps: DeepCache~\cite{ma2024deepcache}, $\Delta$-DiT~\cite{chen2024delta}, PAB~\cite{zhao2024pab}, FasterCache~\cite{lv2024fastercache}, and TeaCache~\cite{liu2024timestep} reuse activations or estimate output similarity to skip computation. However, these methods use content-agnostic caching policies, which can be unreliable during novel exploration in interactive world models.

\noindent\textbf{Sparse Attention for Video Generation.} Sparse attention methods for video DiTs~\cite{xi2025sparse, zhang2025fast, zhang2025spargeattn, zhang2025vsa, wu2025vmoba, yang2025sparsevideogen2, huang2025linvideo} exploit spatial-temporal head specialization and achieve 2.28--2.30$\times$ speedups~\cite{xi2025sparse, yang2025sparsevideogen2} on standard bidirectional generation. However, adapting these methods to autoregressive generation remains largely unexplored, as causal constraints and data reordering overhead substantially weaken practical gains without hardware-aware kernel design.

\begin{figure*}[!t]
    \centering
    \includegraphics[width=\textwidth]{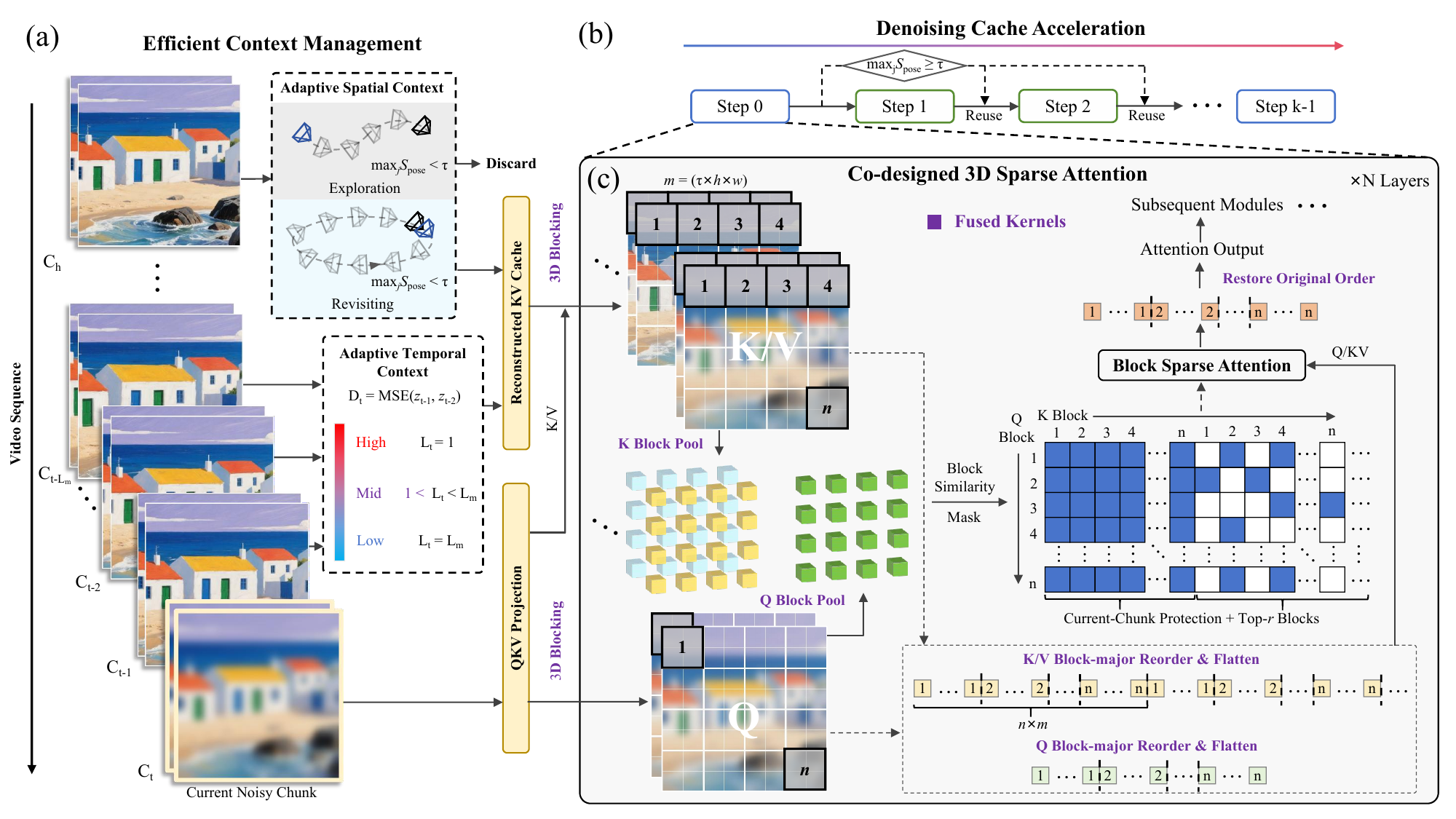}

    \caption{Overview of Light Interaction. (a) Adaptive context management selects valid temporal context and retrieved spatial memory to reconstruct the KV cache for the current chunk. (b) Denoising cache acceleration reuses early-step model outputs for intermediate denoising steps during revisiting, while preserving normal computation at the first step and the final correction step. (c) Co-designed 3D sparse attention partitions the reconstructed KV cache and current queries into 3D blocks, computes block-level similarity from pooled block representations to form a sparse mask, and executes sparse attention with fused kernels for query preparation, KV preparation, and layout restoration.}
    \label{fig:light_interaction}
    \vspace{-1em}
\end{figure*}

\section{Light Interaction}

As illustrated in Figure~\ref{fig:light_interaction}, Light Interaction combines trajectory-gated computation reduction with an AR-aware sparse attention backend: (a)~\textbf{Adaptive Context Management} gates retrieved spatial memory by camera-pose-aware similarity and adapts temporal context according to local latent dynamics; (b)~\textbf{Denoising Cache Acceleration} reuses early-step model outputs only when camera-pose-aware similarity indicates reliable revisiting; and (c)~\textbf{Hardware-Software Co-designed 3D Block Sparse Attention} makes the remaining historical attention efficient under causal AR constraints.

\subsection{Adaptive Context Management}

In autoregressive interactive video generation, contextual history is essential for suppressing error accumulation and maintaining long-horizon coherence. In practice, it mainly takes two complementary forms: \emph{temporal context} and \emph{retrieved spatial memory}. Temporal context refers to recent local history along the generation trajectory and supports short-range motion continuity. Retrieved spatial memory refers to long-range history retrieved by camera-pose-aware similarity and supports geometric consistency when the camera revisits previously seen regions.

However, spatial memory is useful only when it is geometrically relevant, and the optimal temporal window depends on local scene dynamics. Existing interactive autoregressive video generation models, including HY-WorldPlay~\cite{sun2025worldplay} and Matrix-Game-3.0~\cite{li2026matrixgame3}, typically use fixed-length temporal or spatial memory, which can be suboptimal under dynamic scene changes. We therefore propose a dynamic context management strategy that adaptively selects both.

\paragraph{Temporal Context Adaptive Mechanism.}
Temporal context is selected from recent local history before context reconstruction. Let $L_t$ denote the number of recent historical units retained for temporal conditioning, where a unit can be a frame or a chunk depending on the model. In our implementation, temporal selection is performed at the chunk level before context reconstruction.

Directly comparing the current unit is unreliable because it is still in a noisy pre-denoising state and does not provide a reliable reference for temporal validity. Instead, let $z_t$ denote the latent representation of the $t$-th historical unit, and we estimate local dynamics from the two most recent stable historical units in latent space:
\begin{equation}
D_t = \mathrm{MSE}(z_{t-1}, z_{t-2}),
\end{equation}
where $\mathrm{MSE}(\cdot,\cdot)$ is averaged over all latent dimensions. To reduce short-term oscillation, we smooth the instantaneous dynamics with an exponential moving average:
\begin{equation}
\bar{D}_t = \alpha D_t + (1-\alpha)\bar{D}_{t-1}, \qquad \alpha \in (0,1],
\end{equation}
where $\alpha$ is the smoothing factor and $\bar{D}_t$ is initialized by the first valid $D_t$.

Based on the smoothed dynamics $\bar{D}_t$, we adapt the temporal window within the budget $L_m$:
\begin{equation}
L_t=\mathrm{clip}\!\left(\left\lfloor L_m \cdot \frac{\kappa}{\bar{D}_t+\kappa}\right\rfloor,\; 1,\; L_m\right),
\end{equation}
where $\kappa>0$ is on the scale of $\bar{D}_t$ and controls the sensitivity. This shrinks the temporal window under large dynamics and expands it under stable dynamics.

\paragraph{Retrieved Spatial Memory Adaptive Mechanism.}
Retrieved spatial memory is selected from long-range historical memory according to camera-pose-aware similarity. Let $S_{\text{pose}}(t,j)$ denote the pose-aware similarity between the current view at time $t$ and the $j$-th retrieved historical candidate, where larger values indicate higher geometric relevance. During revisiting, such context provides useful conditioning from geometrically relevant past views. During exploration, however, pose-aware retrieval may still return the most similar historical candidate even when no valid match exists, introducing irrelevant context and redundant downstream computation.

To prevent forced retrieval, we define an absolute pose-similarity threshold $\tau_{\text{pose}}$. When the maximum retrieval similarity satisfies
\begin{equation}
\max_j S_{\text{pose}}(t,j) < \tau_{\text{pose}},
\end{equation}
the current state is identified as a \emph{pure exploration} phase. In this case, the retrieved spatial memory is discarded and excluded from subsequent context reconstruction. Otherwise, the matched retrieved spatial memory is retained for conditioning. In both HY-WorldPlay and Matrix-Game-3.0, we instantiate $S_{\text{pose}}$ with $S_{\text{FOV}}$, and $\tau_{\text{pose}}$ with $\tau_{\text{FOV}}$. This mechanism ensures that only valid long-range spatial memory is incorporated, while reducing the effective context length in unseen regions.

\subsection{Lightweight Denoising Cache Acceleration}

Rectified-Flow-based interactive video generators are typically executed with very few denoising steps after distillation~\cite{lee2024improving}. Under such a short denoising horizon, adjacent model evaluations can be partially redundant, but this redundancy is highly state-dependent. During exploration of unseen regions, generation is only weakly anchored by historical context, leading to larger step-to-step variation in the denoising trajectory. In contrast, during revisiting, reliable spatial memory provides stronger geometric constraints, resulting in a more stable denoising process.

\begin{figure}[t]
    \centering
    \includegraphics[width=0.6\linewidth]{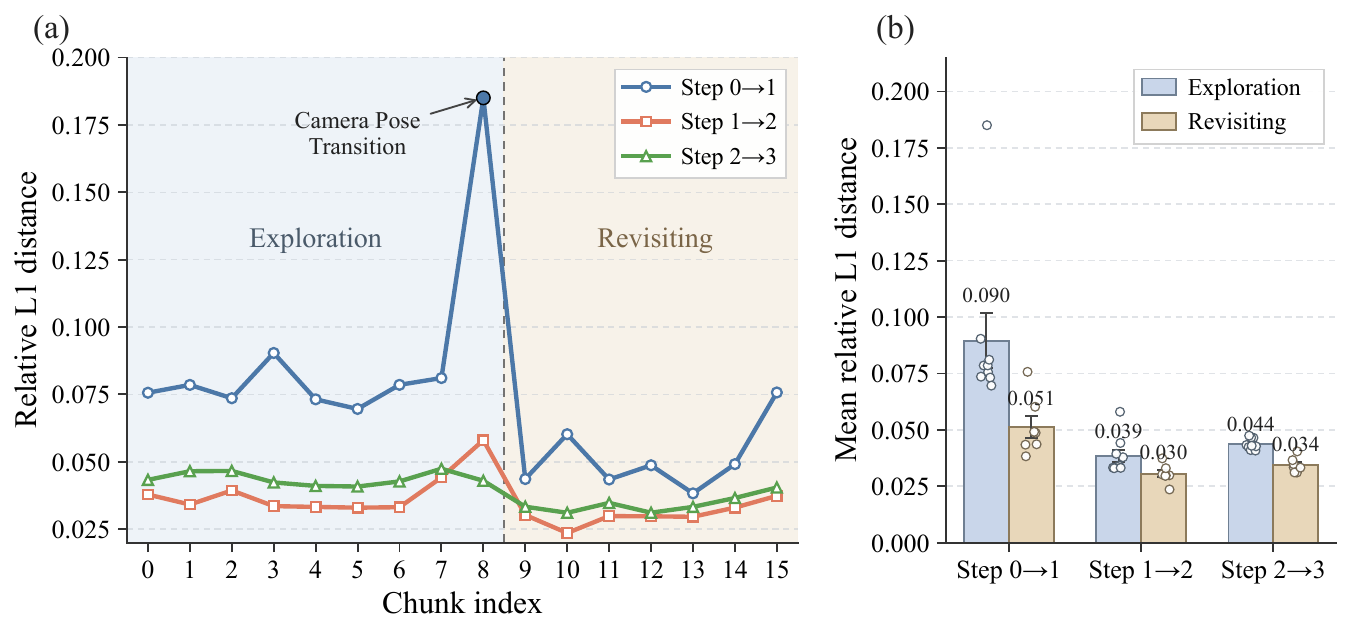}
    \vspace{-1em}
    \caption{Relative L1 distances of consecutive denoising-step pairs in exploration and revisiting, where $\mathrm{RelL1}(y_{s-1},y_s)=\|y_s-y_{s-1}\|_1/(\|y_{s-1}\|_1+\epsilon)$ and $\epsilon=10^{-8}$ is a small constant for numerical stability. (a) Chunk-wise relative L1 distance for Step $0\!\to\!1$, Step $1\!\to\!2$, and Step $2\!\to\!3$. (b) Mean relative L1 distance for each step pair in the two phases, with chunk-level samples overlaid.}
    \label{fig:denoise_diff}
    \vspace{-1em}
\end{figure}

As shown in Figure~\ref{fig:denoise_diff}, adjacent-step discrepancies are lower during revisiting than during exploration. Motivated by this observation, we enable denoising cache reuse only when the current view has a reliable pose-aware historical reference.

Specifically, we reuse the same camera-pose-aware signal as in adaptive retrieved spatial memory selection, and activate denoising cache acceleration only when
\begin{equation}
\max_j S_{\text{pose}}(t,j) \geq \tau_{\text{pose}}.
\end{equation}
In both HY-WorldPlay and Matrix-Game-3.0, this condition is instantiated as
\begin{equation}
\max_j S_{\text{FOV}}(t,j) \geq \tau_{\text{FOV}}.
\end{equation}
When this condition is satisfied, we evaluate the model at the first denoising step, and reuse its output to approximate the intermediate denoising steps. Let
\begin{equation}
v_{\theta}(x_i,t_i,c)
\end{equation}
denote the model output at step $i$, where $c$ includes the selected context and camera conditioning. For a $K$-step denoising process, after obtaining the first-step output $v_{\theta}(x_0,t_0,c)$, we reuse it for all intermediate steps
$i \in \{1,\dots,K-2\}$:
\begin{equation}
v_{\theta}(x_i,t_i,c) \approx v_{\theta}(x_0,t_0,c), \qquad i \in \{1,\dots,K-2\}.
\end{equation}
The model is called only at the first and final denoising steps, while other steps reuse the first-step output. When $K \leq 2$, there are no intermediate steps to approximate, and no reuse is applied.

The final denoising step is always computed normally to correct accumulated deviations before decoding. This design reduces repeated Transformer evaluations while restricting reuse to revisiting regimes where it is empirically more reliable.

\subsection{Hardware-Software Co-designed 3D Sparse Attention}

Existing sparse video attention methods are not directly suitable for interactive autoregressive video generation. SVG sparse patterns~\cite{xi2025sparse} are mainly designed for non-autoregressive settings, while LongCat-Video-style 3D block sparsity~\cite{team2025longcat} still suffers from substantial memory overhead caused by block gathering, layout conversion, and scattered memory access. We therefore adapt 3D block sparse attention to the autoregressive setting and further optimize its execution with fused operators.

\paragraph{Autoregressive Adaptation of 3D Block Sparse Attention.}

3D block sparse attention organizes visual tokens into regular spatiotemporal blocks and performs block-level selection. Unlike token-level pruning, this preserves the local structure of video data. In an autoregressive configuration, the model retains all text-conditioning tokens and current-frame denoising tokens. Sparsification is applied only to the historical visual KV cache, where tokens are partitioned into non-overlapping 3D blocks of size $(B_t, B_h, B_w)$. For each attention head, block pooling and sparse selection are performed independently.

For each query block $\mathcal{Q}_i$, we derive a pooled proxy vector to calculate relevance scores for the historical visual KV blocks:
\begin{equation}
\bar{q}_i=
\frac{1}{|\mathcal{Q}_i|}
\sum_{j\in\mathcal{Q}_i} q_j,
\qquad
\bar{k}_m=
\frac{1}{|\mathcal{K}_m|}
\sum_{n\in\mathcal{K}_m} k_n.
\end{equation}
Let $M$ denote the number of historical visual KV blocks, and let $r\in(0,1]$ denote the retained fraction. We then select the retained historical block indices as
\begin{equation}
\mathcal{I}_i=
\operatorname{Top}_{\lfloor rM\rfloor}
\left\{
\frac{\bar{q}_i^\top \bar{k}_m}{\sqrt{d}}
\right\}_{m=1}^{M},
\end{equation}
where $M$ is the number of historical visual KV blocks and $r\in(0,1]$ denotes the retained fraction. The same indices are then used to gather both K and V blocks. Let $\mathcal{S}_i$ denote the selected historical visual KV blocks induced by $\mathcal{I}_i$. The final attention context for query block $i$ is
\begin{equation}
\mathcal{C}_i
=
\mathcal{T}
\cup
\mathcal{K}^{\text{curr}}
\cup
\mathcal{S}_i,
\end{equation}
where $\mathcal{T}$ denotes all text-condition blocks and $\mathcal{K}^{\text{curr}}$ denotes all KV blocks from the current denoising frame. Therefore, sparsification is restricted to historical visual memory, while text tokens and current-frame tokens remain fully preserved.

\paragraph{Hardware-Aware Operator Fusion.}
We adopt a LongCat-style 3D block sparse attention kernel as the sparse attention core and optimize the surrounding autoregressive dataflow. The sparse pattern alone does not guarantee practical speedup, because block preparation and output restoration are dominated by memory movement. Since tokens in the same 3D block are non-contiguous in the original linear layout, a naive implementation would require separate operators for block gathering, mean pooling, layout conversion, boundary padding, and output scattering. We therefore fuse the sparse dataflow into three Triton kernels:

\begin{itemize}[leftmargin=*]
    \item \textbf{Fused Q-Preparation:} This kernel fuses query block tiling, block-wise mean pooling, and block-major layout generation. For each query block, it maps spatiotemporal block coordinates to linear token indices, writes query features to a contiguous block-major buffer, and simultaneously computes the pooled query feature for sparse index generation. Boundary cases are handled by masked loads, avoiding separate padding or copy operations.

    \item \textbf{Fused KV-Preparation:} This kernel reads K and V jointly from the visual KV input, performs 3D block tiling and block-major layout conversion, and writes tiled K/V blocks into a global block-major KV buffer. During write-back, pointer offsets skip the preallocated text-token region, so text tokens are preserved while visual blocks are appended contiguously without extra concatenation. The same pass also computes pooled K features for block-level similarity scoring.

    \item \textbf{Fused Untile Scatter:} After sparse attention on the block-major layout, this kernel restores the output to the original autoregressive linear token order. It maps each output block back to its temporal-spatial coordinates and writes valid tokens into the dense output tensor, while discarding invalid boundary-padding positions through masked stores.
\end{itemize}

Together, these fused operators eliminate redundant intermediate tensors and reduce repeated gather/scatter, layout conversion, and padding overhead, making 3D sparse attention practically effective in autoregressive interactive video generation.

\section{Experiments}
\label{sec:experiments}

\subsection{Experimental Setup}

\textbf{Models.} We evaluate Light Interaction on two state-of-the-art open-source interactive video generation models: \textbf{HY-World1.5-Autoregressive-480P-I2V-distill-8B (HY-WorldPlay)}~\cite{sun2025worldplay} and \textbf{Matrix-Game-3.0-base-distill-5B}~\cite{li2026matrixgame3}. Following HY-WorldPlay, we adopt two predefined camera trajectories, \emph{forward-backward} and \emph{left-right}, and report all main results averaged over both settings.

\textbf{Evaluation Metrics.}
\begin{itemize}[leftmargin=*]
    \item \textit{Quality Metrics.} Following HY-WorldPlay, we report PSNR, SSIM, and LPIPS under both \textit{vs. Original} and \textit{Self-Comparison}. We also report VBench~\cite{huang2024vbench}; following SVG2~\cite{yang2025sparsevideogen2}, we use the averaged VBench score as the final result.
    \item \textit{Efficiency Metrics.} We report latency, speedup ratio, and peak memory consumption. Since VAE decoding introduces a nearly constant overhead across methods, we exclude VAE time from efficiency measurements and report only the latency of the generative backbone.
\end{itemize}

\textbf{Datasets.} We construct the evaluation set from image-text pairs in VBench~\cite{huang2024vbench}. Following recent works such as SVG2, we further refine the original prompts using LLaVA-1.6 to obtain richer descriptions for interactive image-to-video generation. In total, we use 200 image-text pairs.

\textbf{Baselines.} We compare against three representative training-free acceleration baselines: \textbf{Sparse VideoGen (SVG)}~\cite{xi2025sparse}, a static sparse attention method; \textbf{LongCat-Video-BlockSparseAttention (BSA)}~\cite{team2025longcat}, a dynamic 3D block-wise sparse attention baseline adapted from LongCat-Video; and \textbf{TeaCache}~\cite{liu2024timestep}, a denoising cache acceleration method. We follow official configurations and adapt each baseline to the target architecture.

\textbf{Implementation Details.} All experiments are conducted on NVIDIA A100 (80GB) GPUs. For sparse methods, we match the sparse computation volume within each model. In HY-WorldPlay, our method retains 17.5\% of historical KV cache tokens for attention computation, while BSA uses a global retained ratio of 31.25\%; under the longest context, these settings yield matched sparse computation volume. SVG uses \texttt{mul\_val}=2 on both models, and TeaCache uses $\delta=0.1$. The camera-pose similarity threshold is set to 0.7 on HY-WorldPlay and 0.45 on Matrix-Game-3.0. Unless otherwise specified, all quantitative results are averaged over the full evaluation set.

\subsection{Overall Performance Evaluation}

Table~\ref{tab:performance_comparison} reports the quantitative comparison on HY-WorldPlay and Matrix-Game-3.0. Overall, Light Interaction achieves the best quality-efficiency trade-off on HY-WorldPlay and the fastest runtime on Matrix-Game-3.0 with competitive visual quality.

On \textbf{HY-WorldPlay}, our method achieves the strongest overall performance among all baselines, providing the best fidelity to the original model, the best self-comparison consistency, a $2.59\times$ speedup, 140.36\,s lower latency, and 21.91\,GB less peak memory. SVG and BSA also incur additional adaptation overhead in this autoregressive setting: SVG introduces extra padding under mismatched Q/K lengths, while BSA triggers model offloading due to memory overflow, both of which limit practical acceleration.

On \textbf{Matrix-Game-3.0}, our method achieves the best runtime with a $1.61\times$ speedup. Although TeaCache obtains stronger self-comparison metrics, its lower VBench score suggests that higher retrospective similarity does not necessarily imply better overall perceptual quality. In contrast, our method achieves the fastest runtime while maintaining competitive quality. SVG and BSA also show limited acceleration on this model, suggesting that their sparse patterns are less aligned with its execution characteristics.

\definecolor{OursColor}{RGB}{235,245,255}

\begin{table*}[t]
    \centering
    \caption{Quality and efficiency comparison of Light Interaction and baselines.
    \textbf{vs. Original} compares each method with the original full-computation model.
    \textbf{Self-Comparison} compares frame pairs with similar camera poses within the same revisiting trajectory to evaluate consistency.}
    \label{tab:performance_comparison}
    \resizebox{\textwidth}{!}{
    \begin{tabular}{ll ccc ccc c c c c}
        \toprule
        \multirow{2}{*}{\textbf{Model}}
        & \multirow{2}{*}{\textbf{Method}}
        & \multicolumn{3}{c}{\textbf{vs. Original}}
        & \multicolumn{3}{c}{\textbf{Self-Comparison}}
        & \multirow{2}{*}{\textbf{VBench}$\uparrow$}
        & \multirow{2}{*}{\textbf{Latency}$\downarrow$}
        & \multirow{2}{*}{\textbf{Speedup}$\uparrow$}
        & \multirow{2}{*}{\textbf{Mem.}$\downarrow$} \\
        \cmidrule(lr){3-5} \cmidrule(lr){6-8}
        & & PSNR$\uparrow$ & SSIM$\uparrow$ & LPIPS$\downarrow$
          & PSNR$\uparrow$ & SSIM$\uparrow$ & LPIPS$\downarrow$
          & & (s) & & (GB) \\
        \midrule

        \multirow{5}{*}{\textbf{HY-WorldPlay}}
        & Original  & --    & --     & --     & 18.60 & 0.5678 & 0.2051 & 0.8190 & 228.60 & $1.00\times$ & 76.57 \\
        \cmidrule(lr){2-12}
        & SVG       & 19.48 & 0.6028 & 0.2209 & 17.75 & 0.5299 & 0.2187 & 0.8082 & 247.65 & $0.92\times$ & 77.86 \\
        & BSA       & 15.94 & 0.4639 & 0.3755 & 15.44 & 0.4205 & 0.3720 & 0.7943 & 474.57 & $0.48\times$ & 75.03 \\
        & TeaCache  & 20.90 & \textbf{0.6588} & 0.1892 & \textbf{18.86} & 0.5743 & 0.2054 & 0.8150 & 203.25 & $1.12\times$ & 76.64 \\
        \rowcolor{OursColor}
        & \textbf{Ours} & \textbf{24.81} & 0.6500 & \textbf{0.1788} & 18.85 & \textbf{0.5854} & \textbf{0.1963} & \textbf{0.8220} & \textbf{88.24} & \textbf{2.59$\times$} & \textbf{54.66} \\

        \midrule
        \multirow{5}{*}{\textbf{Matrix-Game-3.0}}
        & Original  & --    & --     & --     & 15.49 & 0.4685 & 0.4048 & 0.7432 & 59.70 & $1.00\times$ & 35.04 \\
        \cmidrule(lr){2-12}
        & SVG       & 12.98 & 0.4170 & 0.5587 & 14.48 & 0.4949 & 0.4406 & \textbf{0.7511} & 96.16 & $0.62\times$ & \textbf{35.02} \\
        & BSA       & 13.34 & 0.4228 & 0.5795 & 16.66 & 0.5326 & 0.4094 & 0.7336 & 63.26 & $0.94\times$ & 35.03 \\
        & TeaCache  & \textbf{19.03} & \textbf{0.5619} & 0.3818 & \textbf{18.84} & \textbf{0.5765} & \textbf{0.3602} & 0.7146 & 41.49 & $1.44\times$ & 35.32 \\
        \rowcolor{OursColor}
        & \textbf{Ours} & 17.76 & 0.5306 & \textbf{0.3692} & 14.63 & 0.4570 & 0.4424 & 0.7350 & \textbf{37.07} & \textbf{1.61$\times$} & 35.04 \\
        \bottomrule
    \end{tabular}
    }
\end{table*}

\subsection{Effectiveness of Individual Components}

Table~\ref{tab:component_worldplay} compares the original model, the full Light Interaction method, and variants with only one component enabled on HY-WorldPlay. The results show that the three components are complementary.

\textbf{Context Management.} Temporal context management contributes most to latency and memory reduction, while spatial context management contributes more to fidelity to the original model. Combining them yields the strongest standalone quality gain and the best self-comparison score.

\textbf{Denoising Cache Acceleration.} Denoising cache provides additional speedup by reducing the effective denoising cost from 4 steps to about 3 steps on average. It also achieves the highest PSNR against the original model, because the proposed dynamic denoising mechanism activates reuse only when intermediate-step approximation is expected to introduce limited error.

\textbf{3D Sparse Attention.} 3D sparse attention is a major source of acceleration. Although it causes some quality degradation when used alone, this effect is largely compensated when combined with context management, which improves the quality of the retained context before sparse execution.

\begin{figure}[t]
    \centering
    \begin{minipage}[t]{0.48\columnwidth}
        \centering
        \includegraphics[width=\linewidth]{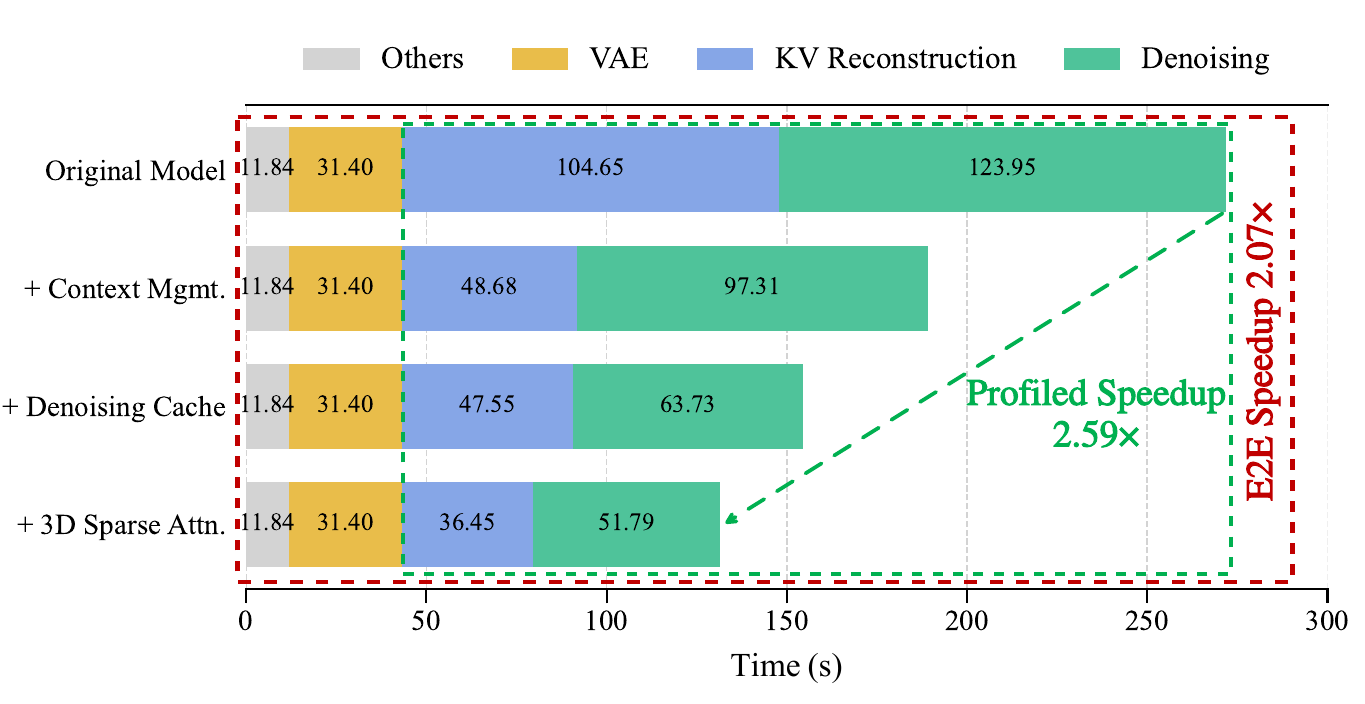}
        \captionof{figure}{Stage-wise latency breakdown on HY-WorldPlay under progressive module enabling.}
        \label{fig:module_latency_breakdown}
    \end{minipage}
    \hfill
    \begin{minipage}[t]{0.48\columnwidth}
        \centering
        \includegraphics[width=\linewidth]{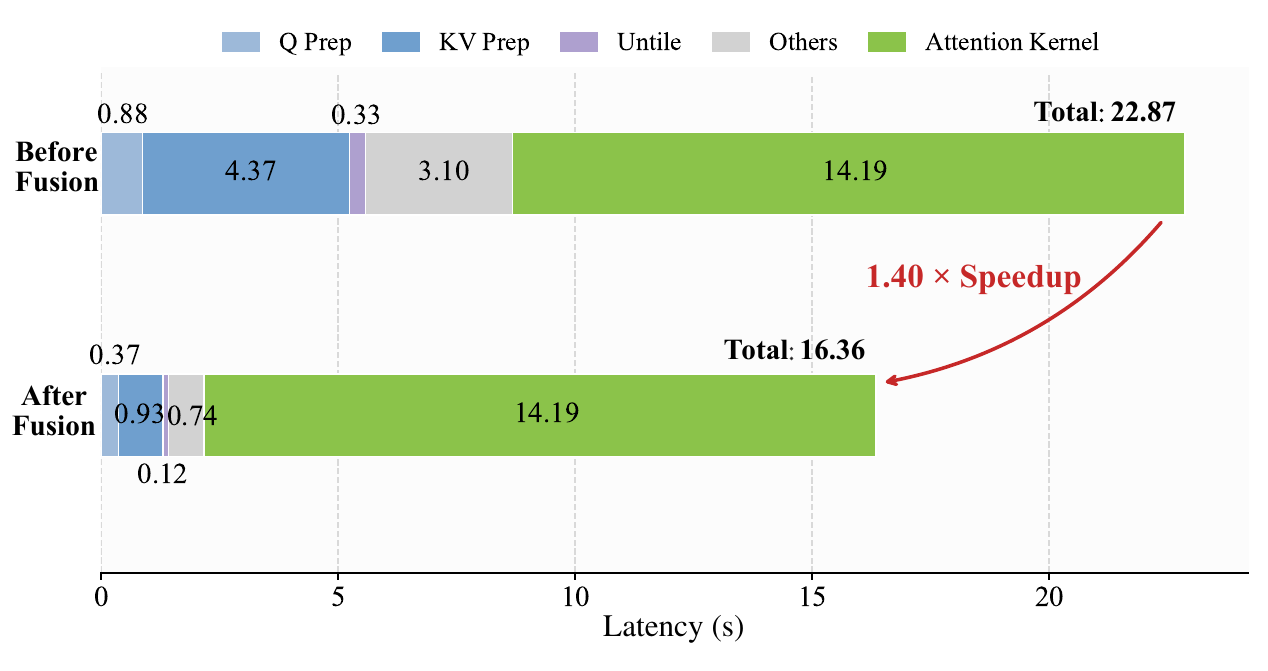}
        \captionof{figure}{Latency of core sparse operators on HY-WorldPlay before and after kernel fusion.}
        \label{fig:fused_kernel_efficiency}
    \end{minipage}

\end{figure}

Figure~\ref{fig:module_latency_breakdown} shows the stage-wise latency breakdown as modules are progressively enabled. Context Management mainly reduces KV reconstruction cost, Denoising Cache Acceleration shortens the denoising stage, and 3D Sparse Attention further lowers the remaining generation cost.

Figure~\ref{fig:fused_kernel_efficiency} further shows that kernel fusion reduces surrounding operator overhead without changing the sparse attention kernel itself, with the largest gain from \textit{KV Prep}. Overall, kernel fusion brings a 1.40$\times$ speedup to the sparse-attention portion.

\definecolor{LightBlue}{RGB}{235,245,255}
\definecolor{LightGray}{RGB}{248,248,248}

\begin{table}[t]
    \centering
    \caption{Effectiveness of individual components of Light Interaction on HY-WorldPlay.}
    \label{tab:component_worldplay}
    \resizebox{0.8\columnwidth}{!}{
    \begin{tabular}{lcccccc}
        \toprule
        \rowcolor{LightGray}
        \textbf{Variant}
        & \textbf{Latency}$\downarrow$
        & \textbf{Speedup}$\uparrow$
        & \textbf{vs. Orig.}
        & \textbf{Self-Comp.}
        & \textbf{VBench}$\uparrow$
        & \textbf{Mem.}$\downarrow$ \\

        \rowcolor{LightGray}
        & \textbf{(s)}
        &
        & \textbf{PSNR}$\uparrow$
        & \textbf{PSNR}$\uparrow$
        &
        & \textbf{(GB)} \\
        \midrule

        Original Model
        & 228.60 & $1.00\times$ & --
        & 18.60 & 0.8190 & 76.57 \\

        \midrule

        Only Context Mgmt. (Temporal)
        & 152.88 & $1.50\times$ & 31.98
        & 20.11 & 0.8208 & \textbf{54.66} \\

        Only Context Mgmt. (Spatial)
        & 213.71 & $1.07\times$ & 38.73
        & 18.85 & 0.8191 & 76.57 \\

        Only Context Mgmt. (Full)
        & 144.49 & $1.58\times$ & 29.24
        & \textbf{20.20} & 0.8210 & \textbf{54.66} \\

        Only Denoising Cache
        & 198.10 & $1.15\times$ & \textbf{55.16}
        & 19.02 & 0.8199 & 76.57 \\

        Only 3D Sparse Attn.
        & 153.69 & $1.49\times$ & 25.53
        & 18.27 & 0.8208 & 76.57 \\

        \midrule

        \rowcolor{LightBlue}
        \textbf{Full Light Interaction}
        & \textbf{88.24}
        & \textbf{2.59$\times$}
        & 24.81
        & 18.85
        & \textbf{0.8220}
        & \textbf{54.66} \\

        \bottomrule
    \end{tabular}
    }

\end{table}

\subsection{Hyperparameter Study}
\label{sec:hyperparameter_study}

\begin{figure*}[t]
    \centering

    \begin{minipage}[t]{0.44\textwidth}
        \centering
        \includegraphics[width=\linewidth]{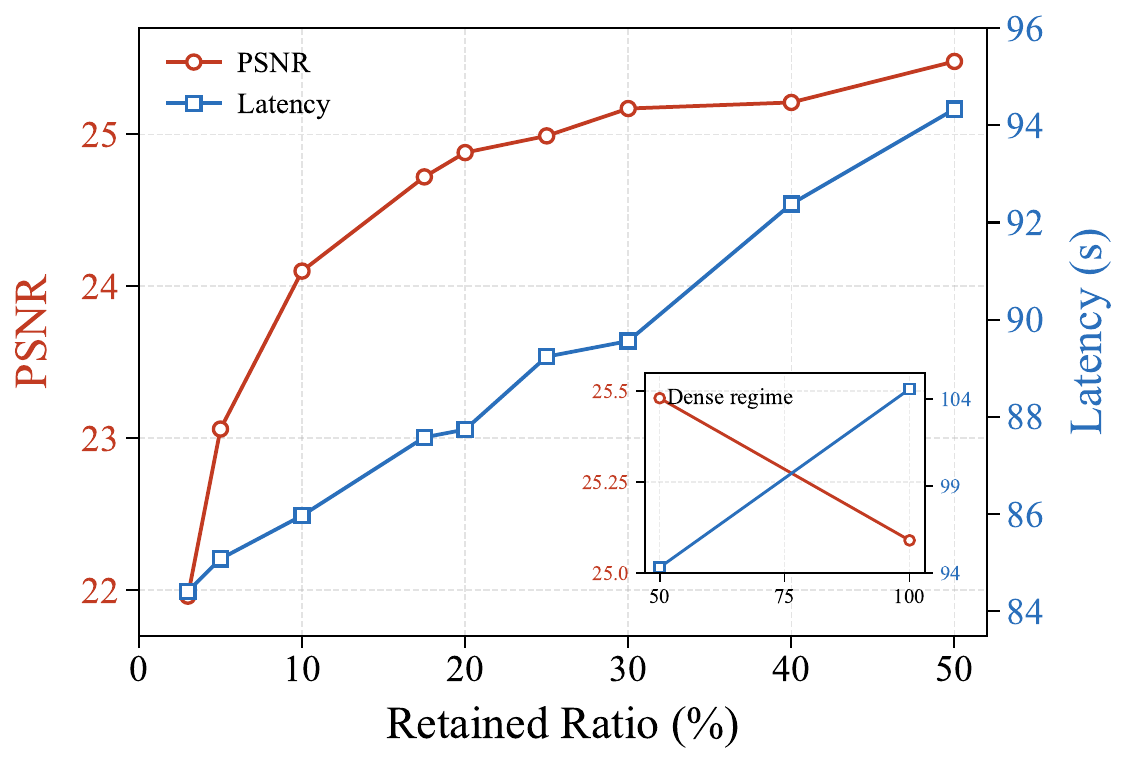}
        \caption{Quality--efficiency trade-off under different retained ratios on HY-WorldPlay.}
        \label{fig:quality_efficiency_tradeoff}
    \end{minipage}
    \hspace{0.01\textwidth}
    \begin{minipage}[t]{0.44\textwidth}
        \centering
        \includegraphics[width=\linewidth]{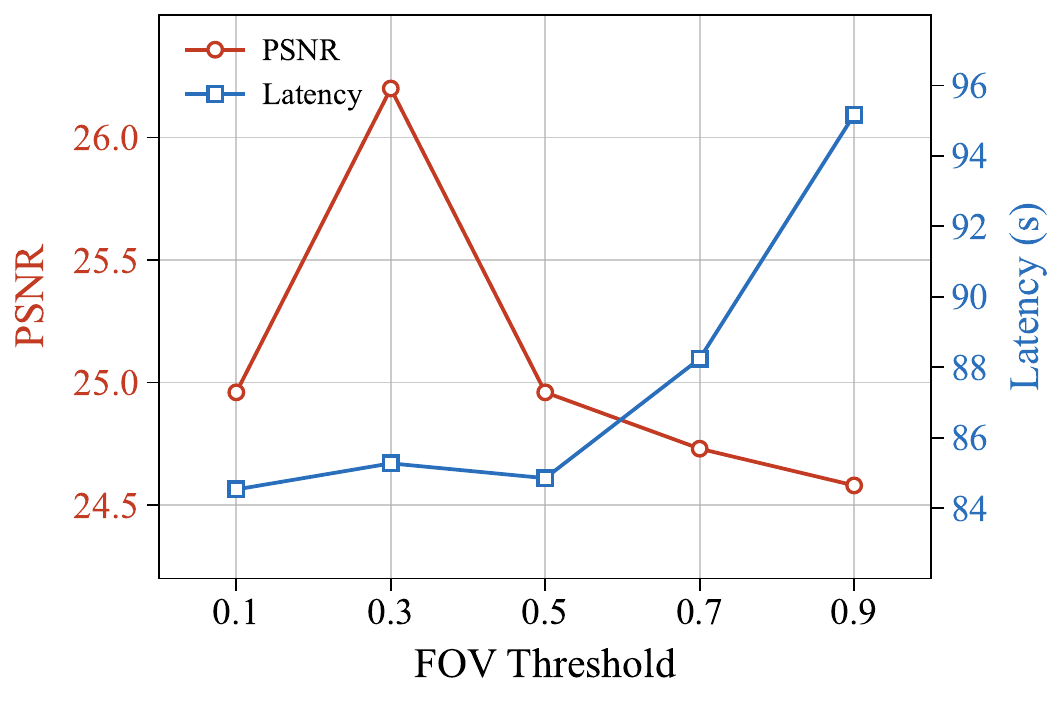}
        \caption{Quality--efficiency trade-off under different camera-pose similarity thresholds on HY-WorldPlay.}
        \label{fig:fov_threshold_analysis}
    \end{minipage}

\end{figure*}

\textbf{Retained Ratio.} Figure~\ref{fig:quality_efficiency_tradeoff} shows the effect of the retained ratio in sparse context selection on HY-WorldPlay. Increasing the retained ratio preserves more historical context and improves reconstruction quality, but also weakens the runtime advantage of sparsity. The adopted sparse setting provides a balanced operating point between quality and efficiency.

\textbf{Camera-Pose Similarity Threshold.} Figure~\ref{fig:fov_threshold_analysis} shows the effect of the camera-pose similarity threshold used for adaptive gating. Small thresholds trigger spatial-memory retention and denoising-cache reuse more frequently, while overly large thresholds become too conservative and suppress valid revisiting states. Overall, a moderate threshold provides a favorable quality--efficiency trade-off.

\section{Conclusion}
\label{Conclusion}

We presented \textbf{Light Interaction}, a training-free acceleration framework for interactive video world models. By exploiting trajectory-dependent adaptive computing, Light Interaction reduces computation in three ways: adaptive context management that gates retrieved spatial memory by geometric validity and adapts the temporal window using local latent dynamics; denoising cache acceleration that reuses early-step model outputs to approximate intermediate denoising steps during revisiting; and hardware-software co-designed 3D block sparse attention with Triton fused kernels. Evaluated on HY-WorldPlay and Matrix-Game-3.0, we achieve up to $2.59\times$ speedup without model retraining.

\textbf{Limitations.}
Our framework assumes a camera-pose-aware relevance signal, which must be approximated from camera extrinsics or other geometric cues when not explicitly available.
The denoising-output cache is validated only on short-step denoising models ($K \leq 4$).
Moreover, the realized speedup of the sparse attention backend depends on the memory organization and execution structure of the underlying autoregressive interactive video model.

\section*{Broader Impacts}

Light Interaction accelerates interactive world models, improving accessibility for research and paving the way toward more responsive applications in embodied AI, game simulation, and virtual scene navigation. The primary societal risk is that faster generation lowers the barrier to creating synthetic media at scale; however, our method is inference-only and does not expand the model capabilities. Detection and attribution tools should continue to evolve alongside efficiency research.

\small
\bibliographystyle{unsrtnat}
\bibliography{ref}

\clearpage
\appendix
\normalsize

\section{Leave-One-Out Ablation of Light Interaction on HY-WorldPlay}
\label{appendix_ablation_worldplay}

We perform a leave-one-out ablation on HY-WorldPlay using a fixed subset for computationally intensive analysis. Unlike Table~\ref{tab:component_worldplay}, which measures the standalone contribution of each module, this experiment starts from the full system and removes one component at a time to test whether each part is necessary within the integrated pipeline.

\begin{table}[H]
    \centering
    \caption{Leave-one-out ablation of Light Interaction on HY-WorldPlay. Due to the high cost of this analysis, results are averaged over a small fixed evaluation subset rather than the full benchmark.}
    \label{tab:ablation_worldplay_appendix}
    \resizebox{\columnwidth}{!}{
    \begin{tabular}{lcccccc}
        \toprule
        \textbf{Variant} & \textbf{Latency (s)}$\downarrow$ & \textbf{Speedup}$\uparrow$ & \textbf{PSNR}$\uparrow$ & \textbf{Self-Comp. PSNR}$\uparrow$ & \textbf{VBench}$\uparrow$ & \textbf{Peak Mem. (GB)}$\downarrow$ \\
        \midrule
        w/o 3D Sparse Attn.    & 110.04 & 2.08$\times$ & \textbf{26.77} & \textbf{19.99} & 0.8285 & \textbf{54.66} \\
        w/o KV Cache Mgmt.     & 133.19 & 1.72$\times$ & 25.43 & 18.58 & \textbf{0.8329} & 76.57 \\
        w/o Denoising Cache    & 111.28 & 2.05$\times$ & 25.02 & 19.27 & 0.8314 & \textbf{54.66} \\
        Full Light Interaction & \textbf{88.24} & \textbf{2.59$\times$} & 24.72 & 18.51 & 0.8295 & \textbf{54.66} \\
        \bottomrule
    \end{tabular}
    }
\end{table}

Table~\ref{tab:ablation_worldplay_appendix} shows that removing any single component weakens the overall trade-off, although the failure mode differs across modules. Removing \textbf{3D Sparse Attention} yields the strongest quality recovery, which is expected because denser attention preserves more information, but it also incurs a substantial latency increase. Removing \textbf{KV Cache Management} degrades both speed and memory efficiency, confirming that controlling historical context growth is a core requirement rather than a secondary refinement. Removing \textbf{Denoising Cache} also increases runtime, but the degradation is smaller, indicating that this module serves as a lightweight complementary accelerator on top of the other two components.

\section{Additional Implementation Details}
\label{appendix_impl_details}

\paragraph{Model-specific instantiation.}
The temporal context formulation in Section~3.1 is presented in a general form to describe a broader adaptive mechanism. In the current experiments, HY-WorldPlay uses a simplified instantiation that retains only the most recent temporal unit (i.e., $L_t=1$), since local dynamics in this model are typically strong. Matrix-Game-3.0 does not enable the parameterized temporal-window adaptation in the current implementation. The camera-pose similarity threshold and sparse retained ratio follow the settings in Section~4.

\paragraph{Sparse attention configuration.}
For the hardware-software co-designed 3D sparse attention in Section~3.3, the 3D block size is set to $(4,8,4)$ on HY-WorldPlay and $(4,4,8)$ on Matrix-Game-3.0. Both settings use the same block volume of 128 tokens. In all experiments, sparsification is applied only to the historical visual KV cache, while text tokens and current-frame denoising KV remain fully preserved.

\paragraph{Warm-up behavior.}
To avoid unstable decisions when historical information is still insufficient, we keep the first three chunks in full dense computation without adaptive pruning or denoising cache reuse. The adaptive mechanisms are enabled only after sufficient history has been accumulated.

\paragraph{Overhead accounting.}
The overhead of latent-dynamics estimation and camera-pose/FoV similarity computation is negligible compared with the generative backbone. Sparse index generation is counted as part of sparse attention. In the profiled timing analysis, we focus on the two dominant stages affected by the proposed method: KV reconstruction and denoising computation.

\end{document}